\documentclass[conference]{IEEEtran}

\usepackage{graphicx}
\usepackage{caption}
\graphicspath{
    {figures/}
}

\usepackage{amsmath,amssymb,enumerate}

\usepackage{amsthm}
\usepackage{setspace}
\usepackage{booktabs}
\usepackage[usenames,dvipsnames,svgnames,table]{xcolor}
\usepackage{mathtools}
\usepackage{algorithm, algorithmicx, algpseudocode}
\usepackage{blindtext}
\usepackage{gensymb}
\usepackage{xparse}
\usepackage{lipsum}
\usepackage{mathrsfs}
\usepackage[mathscr]{euscript}
\usepackage{times}
\usepackage[numbers,sort&compress]{natbib}
\usepackage{multicol}
\definecolor{darkgreen}{rgb}{0,0.6,0}
\usepackage[bookmarks=true,colorlinks=true,pdfpagemode=UseNone,citecolor=darkgreen,linkcolor=black,urlcolor=BrickRed,pagebackref]{hyperref}
\usepackage[caption=false,font=footnotesize]{subfig}
\usepackage{amsfonts}
\usepackage{cleveref}
\usepackage[utf8]{inputenc}
\usepackage[T1]{fontenc}
\usepackage{textcomp}
\usepackage{arydshln}
\usepackage{tabu}
\usepackage{cuted}
\usepackage{flushend}

\newtheorem{problem}{Problem}

\newtheorem{remark}{Remark}

\newtheorem{definition}{Definition}

\DeclareMathOperator*{\argmin}{arg\,min}

\renewcommand{\thefigure}{\arabic{figure}}

\definecolor{note}{rgb}{0.1,0.1,1}
\definecolor{rephase}{rgb}{0.15,0.7,0.15}
\definecolor{bag}{rgb}{0.6,0.6,0.2}

\makeatletter
\renewcommand*\env@matrix[1][c]{\hskip -\arraycolsep
  \let\@ifnextchar\new@ifnextchar
  \array{*\c@MaxMatrixCols #1}}
\makeatother

\newcommand{\m}{\mathop{\mathrm{m}}}

\newcommand{\transpose}{\mathsf{T}}

\makeatletter
\newcommand{\mathleft}{\@fleqntrue\@mathmargin0pt}
\newcommand{\mathcenter}{\@fleqnfalse}
\makeatother

\usepackage{bm}
\usepackage{cleveref}

\title{Toward Safety-Aware Informative Motion Planning for Legged Robots} 

\author{Sangli Teng, Yukai Gong, Jessy W. Grizzle, and Maani Ghaffari\\
University of Michigan, Ann Arbor, MI, USA\\
\tt\small \{sanglit, ykgong, grizzle, maanigj\}@umich.edu\\
}

\begin{document}

\maketitle
\thispagestyle{plain}
\pagestyle{plain}

\begin{abstract}
This paper reports on developing an integrated framework for safety-aware informative motion planning suitable for legged robots. The information-gathering planner takes a dense stochastic map of the environment into account, while safety constraints are enforced via Control Barrier Functions (CBFs). The planner is based on the Incrementally-exploring Information Gathering (IIG) algorithm and allows closed-loop kinodynamic node expansion using a Model Predictive Control (MPC) formalism. Robotic exploration and information gathering problems are inherently path-dependent problems. That is, the information collected along a path depends on the state and observation history. As such, motion planning solely based on a modular cost does not lead to suitable plans for exploration. We propose SAFE-IIG, an integrated informative motion planning algorithm that takes into account: 1) a robot's perceptual field of view via a submodular information function computed over a stochastic map of the environment, 2) a robot's dynamics and safety constraints via discrete-time CBFs and MPC for closed-loop multi-horizon node expansions, and 3) an automatic stopping criterion via setting an information-theoretic planning horizon. The simulation results show that SAFE-IIG can plan a safe and dynamically feasible path while exploring a dense map.
\end{abstract}

\IEEEpeerreviewmaketitle

\section{Introduction}
\label{sec:intro}
Information gathering is an important and widely studied task for mobile robots operating in unknown environments~\citep{binney2012branch,levine2013information,binney2013optimizing,RIG,tabib2016computationally,popovic2017online,hitz2017adaptive,IIG,pulido2020kriging}. Legged robots have demonstrated the promising capability of traversing complex terrains~\citep{eth-complex-terrain-navagation,eth-exploring,dai2014whole, kuindersma2016optimization}, and are one of the natural candidates for such tasks. However, existing informative motion planning methods do not consider legged robots' dynamics or safety constraints during the planning. As the legged robot dynamics is hybrid and highly nonlinear~\citep{westervelt2007feedback}, conventional sampling-based planning methods may result in an infeasible or unsafe path plan for legged robot operation. 

Multi-step stability and feasibility are among the main challenges in developing sampling-based motion planning methods suitable for legged robots. The sampling-based informative motion planning methods~\citep{IIG, RIG} have been studied to plan a path that maximizes the information robot gathered along the path and are asymptotically optimal. However, these algorithms do not encode any safety criteria nor consider legged robot dynamics. The Control Barrier Function (CBF)~\citep{CBF-TAC, ECBF-ACC} has gained success in legged robot control~\citep{ETH-CBF-multi-layer, rssDCBF,Choi-RSS-20} and has been integrated into the sampling-based motion planning framework~\citep{quanCBF-RRT,opt-CBF-RRT} to avoid collisions. \citet{opt-CBF-RRT} integrated CBFs into the Rapidly-exploring Random Tree (RRT) \cite{RRT}. However, the node expansion is one-step ahead, i.e., greedy, using a Quadratic Programming (QP); furthermore, the development of CBFs for legged robot dynamics is not explored. \citet{rssDCBF} developed the Discrete-time Control Barrier Function (DCBF) applied to the bipedal robot navigation. \citet{Bike-MPC-DCBF} combined DCBF with Model Predictive Control (MPC), enabling safe receding horizon control, which could potentially bridge the robot dynamics and safety criteria in a longer time.  


\begin{figure}
    \centering
    \includegraphics[width=\columnwidth]{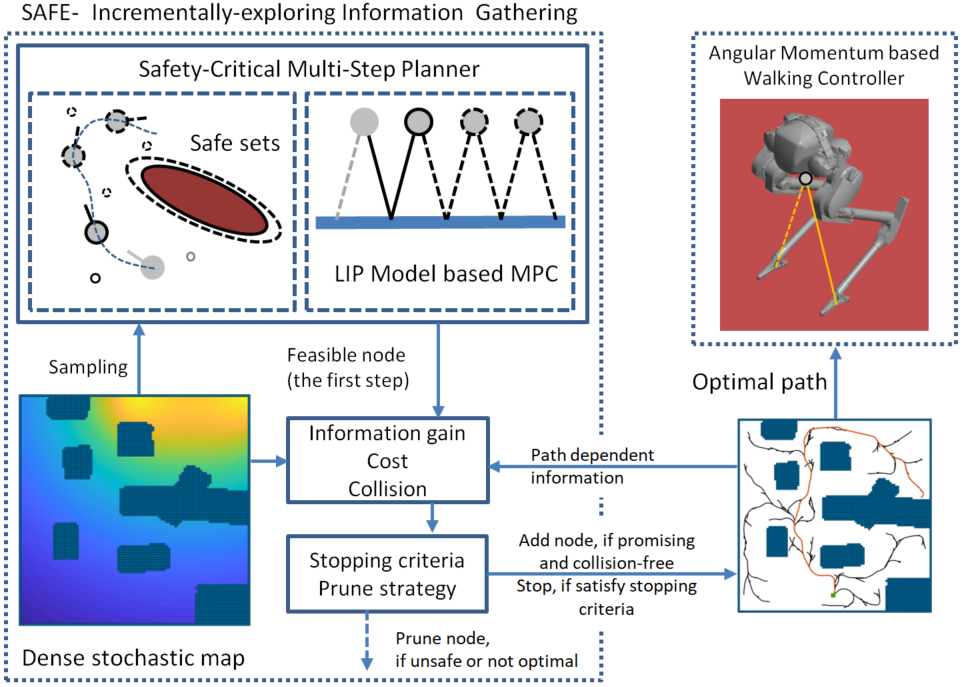}
    \caption{SAFE-IIG framework. Safety-critical multi-step planner plans a multi-step stable path via the LIP model. CBF constraints will guarantee the robot state stays in the forward invariant safe set. The planner performs in a receding horizon manner, such that the first step in each sampling is used for node expansion. A node is added to the path if it is within the dynamic budget; it has promising information gain and is collision-free. The planner automatically stops when an information-theoretic stopping criterion is satisfied. The angular momentum-based walking controller then tracks the optimal path.}
    \label{fig:framework}
\end{figure}

In this paper, to allow a legged robot to navigate with safety and stability while maximizing the information gathered, we propose an informative robot motion planning framework that is tightly coupled with robot dynamics and encodes safety criteria. In particular, our work combines the Incrementally-exploring Information Gathering (IIG)~\cite{IIG} framework for exploration with DCBF-MPC~\cite{rssDCBF,Bike-MPC-DCBF}. Node expansion is enabled by trajectory optimization through DCBF-MPC based on a simplified robot model in a receding horizon manner. 

The main contributions of this paper are as follows.
\begin{itemize}
    \item We develop a DCBF-MPC based safety-critical multi-step planner for legged robot and combine it with RRT for sampling-based motion planning. 
    \item We develop the integration of legged robot dynamics and safety criteria into the sampling-based robotic information gathering algorithms~\citep{RIG,IIG} via the proposed safety-critical multi-step planner. The main result of the proposed planner, SAFE-IIG, enables a robot to navigate an environment safely while collecting information about a dense map. Such scenarios frequently appear in robotic exploration, search and rescue missions, and environmental monitoring problems~\citep{umari2017autonomous,wang2019efficient,rouvcek2019darpa,wang2019autonomous,candela2017planetary,cabrol2018coevolution}.
    \item We validate the planned trajectory's feasibility through simulations of a high degree-of-freedom bipedal robot (a Cassie-series biped robot). The waypoints, velocity, and heading angle computed by the planner are directly used for legged robot path tracking without replanning.  
\end{itemize}
The remainder of this paper is organized as follows. Related work is given in Section~\ref{sec:relatedwork}. We introduce the simplified bipedal robot model and the kinematic constraints for motion planning in Section~\ref{sec:simplebipedmodel}. The DCBF-MPC-based safety-critical multi-step planner is discussed in Section~\ref{sec:multistep-planner}. We develop the proposed safe informative motion planning framework in Section~\ref{sec:safeiig}. Simulation results are presented in Section~\ref{sec:results}. A discussion on the proposed work and its limitations is provided in Section~\ref{sec:discussion}. Finally, Section~\ref{sec:conclusion} concludes the paper and provide future work ideas.  
\section{Related Work}
\label{sec:relatedwork}

The hybrid and nonlinear dynamics of legged robot makes its motion planning difficult~\citep{westervelt2007feedback}. Hybrid Zero Dynamics (HZD)-based periodic gait design \cite{HZD} enables asymptotically stable walking pattern using optimization on the robot full dynamics~\cite{C-FROST, cassie-segway}. The work of \cite{yezhao-phase-space, JHurst-Cassie-MPC, yukaiAngularMomentum} has explored using simplified models, e.g., Linear Inverted Pendulum (LIP) model~\cite{LIPModel} and Spring Loaded Inverted Pendulum (SLIP) model~\cite{SLIP} for bipedal locomotion. To enable multi-step motion planning of Cassie bipedal robots, \citet{JHurst-Cassie-MPC} proposed an MPC framework based on the simplified model~\cite{ACM-simplified}. Robustness to disturbances is achieved by real-time implementation of the MPC at faster than 100 Hz. \citet{yukaiAngularMomentum} combined the LIP model and the angular momentum about the contact point for one-step ahead foot placement planning. The proposed controller makes the robot robust to disturbance as replanning is instantaneous. In this paper, we also adopt the LIP model for the bipedal robot walking task.

CBFs (Control Barrier Functions) \cite{CBF-TAC, ECBF-ACC} have been an effective tool for safety-critical control. They haven been used in stepping stones context for bipedal robot in simulations \cite{quan-3d-stepping-stone, quan-cbf-2d, quan-robust-cbf-2d}. \citet{ETH-CBF-multi-layer} combined MPC and CBF to achieve multi-layer safety in motion generation and control, which is also validated in stepping stone experiment on quadrupedal robot. The discrete-time variant of the CBF, i.e., DCBF is proposed by \citet{rssDCBF}, and implemented via the HZD-based steering controller \cite{robotica-HZD-steering} as a one-step ahead planner for bipedal robot collision avoidance and path tracking. By solving QP for linear obstacles and Quadratically Constrained Quadratic Programming (QCQP) for ellipsoidal obstacles, the robot can track the path while stay in the safe set with minimized steering. All the above CBF-based planning methods are implemented as one-step ahead or focus on stride-to-stride planning. 

Sampling-based motion planning methods are suitable for the exploration of an unknown environment. An example is RRT~\cite{RRT}, which plans a trajectory by sampling the input or configuration space to generate nodes in a tree. For legged robots, randomly sampling the input may enable one-step expansion, but the robot will likely lose stability at future steps. Other methods include Probabilistic Road Map (PRM)~\cite{horsch1994motion,PRM} and their asymptotically optimal variant RRT* and PRM*~\cite{RRT-star}. FIRM~\citep{agha2014firm} provides a probabilistically complete solution to motion planning under sensing uncertainty. The Rapid-exploring Information Gathering (RIG) algorithm \cite{RIG} provides an asymptotically optimal informative motion planning framework. Built on RIG, IIG \cite{IIG} develops information-theoretic convergence criteria and information functions for online implementation. A recent review of asymptotically optimal sampling-based motion planning methods is written by~\citet{gammell4survey}. LQR-Tree \cite{lrq_tree} combines Linear Quadratic Regulator (LQR) and RRT to build a tree of not only trajectories but also a closed-loop controller. However, LQR-Tree can only work for low-dimensional systems due to the complexity of Sum-of-Square programming. We perform trajectory optimization similar to LQR-Tree; however, we adopt a receding horizon approach to expand the trajectory. 

To mitigate the safety problem, CBF is combined with RRT for safe motion planning~\cite{opt-CBF-RRT, quanCBF-RRT}. The work of~\citet{quanCBF-RRT} checks the inequality condition induced by CBF after sampling the input. A node expanded by an input resulting in safety violation is discarded. In \cite{opt-CBF-RRT}, the CBF-QP algorithm is implemented in each sampling step to guarantee safety. Both \cite{opt-CBF-RRT} and \cite{quanCBF-RRT} are formulated for one step node expansion with simple agent dynamics. 
In this work, we adopt DCBF-MPC~\cite{Bike-MPC-DCBF} to optimize the stride-to-stride motion, which will guarantee multi-step stability.

\section{Simplified Bipedal Robot Model}
\label{sec:simplebipedmodel}
\begin{figure}
    \centering
  \includegraphics[width=0.99\columnwidth]{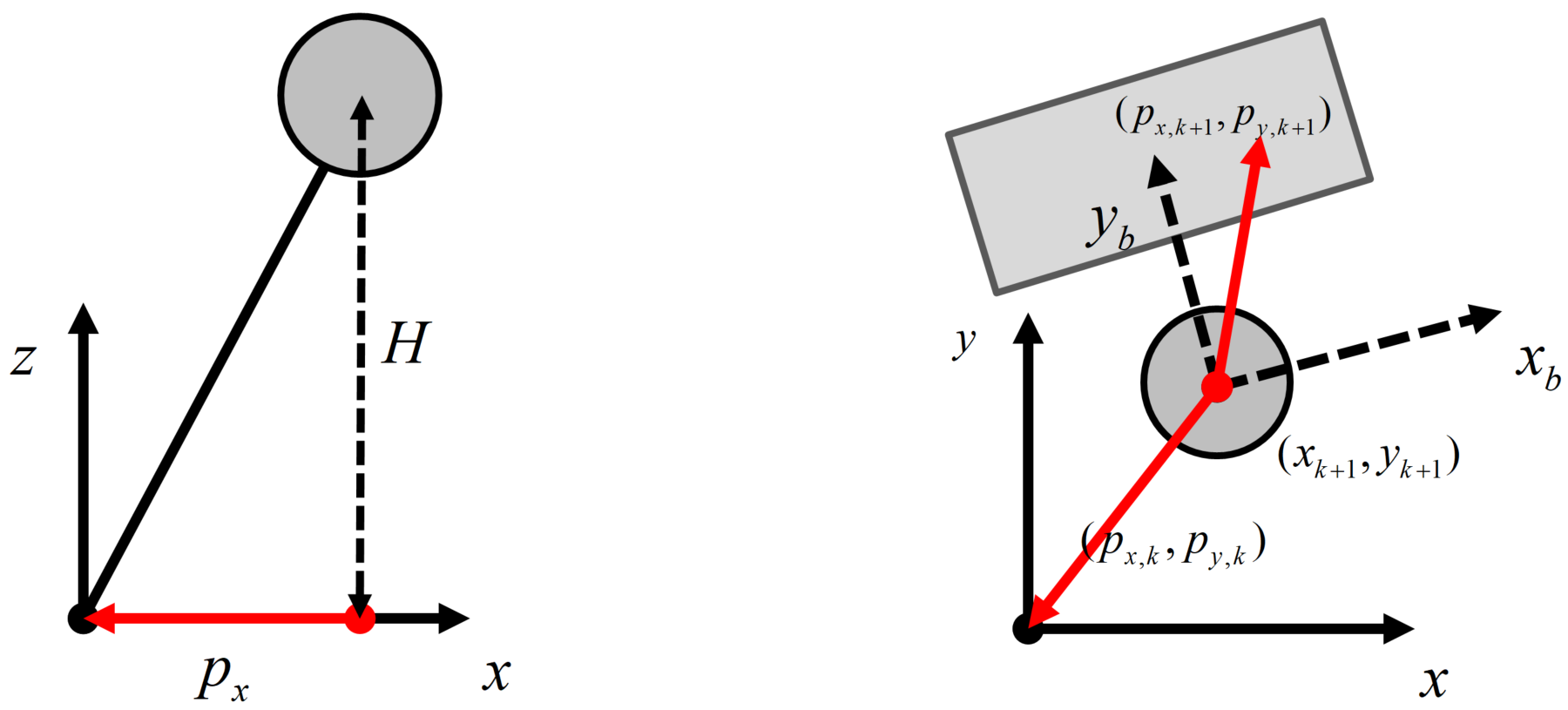}
    \caption{LIP model. Red arrow denotes the vector from robot COM to foot placement. The robot is at the end of step $k$ and about to decide the next foot placement for step $k+1$. The rectangle is the region of reachable foot placement at step $k+1$.}
    \label{fig:lip-model}
\end{figure}

In this section, we introduce the LIP (Linear Inverted Pendulum) model~\cite{LIPModel} and the kinematic constraints for bipedal robot motion planning. We assume the robot is walking on flat ground thus the LIP model is suitable for this application. The LIP model assume constant robot center of mass (CoM) height, massless legs and the stance ankle that is not actuated. 
We assume that the impact is instantaneous and the velocity of robot CoM will not change during the impact. 

Figure~\ref{fig:lip-model} presents the LIP model and the variables. The robot's state variables are the COM position $x$, $y$ and velocity $\dot{x}$ and $\dot{y}$. The input is the desired stance foot position. We use $p_x$ and $p_y$ to denote the distance of the stance foot to the robot's CoM in $x$ and $y$ directions. All the parameters are represented in the world frame. 
\subsection{Linear Inverted Pendulum Model}
We use $H$ to denote the height of COM and $g$ to denote acceleration of gravity. The robot's motion in the $x$ direction within a step satisfies $\ddot{x}=-\frac{g}{H}p_x$, or in state variable form
\begin{equation*}
\begin{aligned}
    \left[ \begin{array}{c}
{\dot{x}} \\ 
{\ddot{x}} \\ 
\end{array} \right]=\left[ \begin{matrix}
   0 & 1  \\
   0 & 0  \\
\end{matrix} \right]\left[ \begin{aligned}
  & x \\ 
 & {\dot{x}} \\ 
\end{aligned} \right]+\left[ \begin{matrix}
   0  \\
   -g/H  \\
\end{matrix} \right]{{p}_{x}}.
\end{aligned}
\end{equation*}
The closed-form solution is given by
\begin{equation*}
\left[ \begin{aligned}
  & x(t) \\ 
 & \dot{x}(t) \\ 
\end{aligned} \right]=\left[ \begin{matrix}
   1 & \sinh (\beta t)/\beta   \\
   0 & \cosh (\beta t)  \\
\end{matrix} \right]\left[ \begin{matrix}
   x(0)  \\
   \dot{x}(0)  \\
\end{matrix} \right]+\left[ \begin{matrix}
   1-\cosh (\beta t)  \\
   -\beta \sinh (\beta t)  \\
\end{matrix} \right]{{p}_{x}},
\end{equation*}
where $\beta =\sqrt{{g}/{H}\;}$.

We assume that the each step has constant time duration $T$. We call the half-open time interval for the $k$-th step, $[kT,(k+1)T)$, as phase $k$. For stride-to-stride control, we use subscript $k$ to denote the state at the \textit{start of phase $k$}.  It follows that the discrete dynamics is given by
\begin{equation}
\label{eq:x_dyn}
\left[ \begin{aligned}
  & {{x}_{k+1}} \\ 
 & {{{\dot{x}}}_{k+1}} \\ 
\end{aligned} \right]=\bm{A}\left[ \begin{aligned}
  & {{x}_{k}} \\ 
 & {{{\dot{x}}}_{k}} \\ 
\end{aligned} \right]+\bm{B}{{p}_{x,k}},
\end{equation}
with
\begin{equation}
\label{eq:AB_mat}
\small{
    \bm{A}:=\left[ \begin{matrix}
   1 & \sinh (\beta T)/\beta   \\
   0 & \cosh (\beta T)  \\
\end{matrix} \right],\bm{B}:=\left[ \begin{matrix}
   1-\cosh (\beta T)  \\
   -\beta \sinh (\beta T)  \\
\end{matrix} \right].
}
\end{equation}
For a LIP model, the motion in the $y$-direction is identical to the $x$-direction, namely
\begin{equation}
\label{eq:y_dyn}
\left[ \begin{aligned}
  & {{y}_{k+1}} \\ 
 & {{{\dot{y}}}_{k+1}} \\ 
\end{aligned} \right]=\bm{A}\left[ \begin{aligned}
  & {{y}_{k}} \\ 
 & {{{\dot{y}}}_{k}} \\ 
\end{aligned} \right]+\bm{B}{{p}_{y,k}}.
\end{equation}

\subsection{Kinematic Constraints} 
We impose reachability constraints on foot placement to avoid violation of joint limits and to decide the contact sequence. The reachable region of the robot's next foot placement is approximated by a square located relative to the CoM at the start of each phase. We use $\theta_k$ to denote the heading angle $\theta$ at the start of phase $k$. The reachable region constraints are given in the robot's body frame as
\begin{equation}
\label{eq:leg_reachable_region}
    \left[ \begin{matrix}
   l{{b}_{x_b,k}}  \\
   l{{b}_{y_b,k}}  \\
\end{matrix} \right]\le {{\left[ \begin{array}{lr}
   \cos ({{\theta }_{k}}) & -\sin ({{\theta }_{k}})  \\
   \sin ({{\theta }_{k}}) & \cos ({{\theta }_{k}})  \\
\end{array} \right]}^{\intercal }}\left[ \begin{matrix}
   {{p}_{x,k}}  \\
   {{p}_{y,k}}  \\
\end{matrix} \right]\le \left[ \begin{matrix}
   u{{b}_{x_b,k}}  \\
   u{{b}_{y_b,k}}  \\
\end{matrix} \right],
\end{equation}
where $ub$ and $lb$ denote the upper and lower bounds of the reachable distances in the robot's longitudinal and lateral directions, which are denoted as $x_b$ and $y_b$, respectively. Because the robot switches its stance leg between phases, $ub$ and $lb$ are time-varying (i.e., they are functions of $k$). Figure~\ref{fig:lip-model} illustrates the reachable region of foot placement relative to the CoM and the robot's body frame.

While taking $\theta$ to be an independent variable can enable more flexible walking patterns, such as diagonal walking, here we constrain it to be consistent with the forward direction of the robot, namely, the heading at time $k$ coincides with the vector from CoM position at step $k$ to $k+1$,
\begin{equation}
\label{eq:heading_angle_constrain}
    \sin ({{\theta }_{k}})=\frac{\Delta {{x}_{k}}}{\sqrt{\Delta {{x}_{k}}^{2}+\Delta {{y}_{k}}^{2}}},\cos ({{\theta }_{k}})=\frac{\Delta {{y}_{k}}}{\sqrt{\Delta {{x}_{k}}^{2}+\Delta {{y}_{k}}^{2}}},
\end{equation}
where $\Delta {{x}_{k}}:={{x}_{k+1}}-{{x}_{k}}$,\ $\Delta {{y}_{k}}:={{y}_{k+1}}-{{y}_{k}}$. By substituting \eqref{eq:heading_angle_constrain} into \eqref{eq:leg_reachable_region}, we can constrain the robot's walking direction.

To avoid infeasible motion, we also impose maximum and minimum CoM position movement in each phase, which are denoted by $l_{\max}$ and $l_{\min}$, respectively,
\begin{equation}
\label{eq:minmax_length}
    {{l}_{\min }}\le \sqrt{\Delta {{x}_{k}}^{2}+\Delta {{y}_{k}}^{2}}\le {{l}_{\max }}.
\end{equation}
\section{Safe Multi-step Planner}
\label{sec:multistep-planner}
In this section, we derive our DCBF-MPC-based safety-critical multi-step planner \cite{Bike-MPC-DCBF}. As a bipedal robot's stride-to-stride motion is discrete, DCBF \cite{rssDCBF} is a natural choice to constrain the robot state in a forward-invariant safe set. However, as the open-loop robot dynamics is unstable and we do not use predefined gait, one step planning may result in failure in later steps. Therefore, MPC is used to ensure multi-step stability. For sampling-based motion planning, multi-step stability is essential to enable feasible node expansion. 
\subsection{Discrete-time Control Barrier Function}
We first introduce the concepts of safety and discrete control barrier functions. For more details, we refer the reader to~\cite{rssDCBF,Bike-MPC-DCBF}.
Consider a discrete-time control system, 
\begin{equation}
 \label{eq:discrete-dynamics}
     \bm x_{k+1} = f(\bm x_{k}, \bm u_{k}),
\end{equation}
where $\bm x_k \in \mathcal{D} \subset \mathbb{R}^{n}$ denotes the system state at time step $k \in \mathbb{Z}^{+}$, $\bm u_k \in \mathcal{U} \subset \mathbb{R}^{m}$ is the control input, and $f$ is a locally Lipschitz function. The safety set $\mathcal{S}$ is defined as the super-level set of a continuously differentiable function $h: \mathbb{R}^{n} \rightarrow \mathbb{R}$,
$\mathcal{S}:=\{\bm x_k \in \mathcal{D} \subset \mathbb{R}^{n} \mid h(\bm x_k) \geq 0\}$. 
Its boundary $\partial \mathcal{S}$ is defined as
$\partial \mathcal{S}: =\{\bm x(k) \in \mathcal{D} \mid h(\bm x_k)=0\}$ and we assume that $\partial \mathcal{S} \cap \partial \mathcal{D}=\varnothing$.
{\color{black}Note that we define $\bm{x}\in \cal D$ that $\cal S \subset \cal D$ to handle disturbances \cite{xu2015robustness, CBF-TAC}.} We wish the robot's states to stay in the safety set once it is in it, which leads to the following definitions of the forward invariance and safety.
\theoremstyle{definition}
\begin{definition}[Forward Invariance and Safety]
A set $\mathcal{S}$ is forward invariant if for every $\bm{x}_0\in\mathcal{S}$, the system trajectory $\bm{x}_k\in\mathcal{S}$ for every $k\ge 0$. The system is safe on $\mathcal{S}$ if set $\mathcal{S}$ is forward invariant~\cite{ETH-CBF-multi-layer}.
\end{definition}
 Before we discuss CBFs, we note that a continuous function $\alpha : (-b,a) \rightarrow (-\infty,\infty)$, for some $a,b\ge 0$ is said to belong to extended class $\mathcal{K}$ ( $\mathcal{K}_e$) if it is strictly increasing and $\alpha(0)=0$. Moreover, if $a,b=\infty$, $\lim _{r \rightarrow \infty} \alpha(r)=\infty$ and $\lim _{r \rightarrow -\infty} \alpha(r)=-\infty$ then $\alpha$ is said to belong to extended class  $\mathcal{K}_{\infty}$ ($\mathcal{K}_{\infty,e}$).


\theoremstyle{definition}
\begin{definition}[Discrete-time Control Barrier Function~\cite{rssDCBF, Bike-MPC-DCBF}]
\label{def:dcbf}
A map $h: {\color{black}\mathcal{D}} \rightarrow \mathbb{R}$ is a Discrete-time Control Barrier Function for \eqref{eq:discrete-dynamics} if there exists a class {\color{black}$\mathcal{K}_{\infty,e}$} function $\alpha$ such that the following hold:
\begin{itemize}
\item [1)] For all $y\in \mathbb{R}_{+}$, $\alpha(y) \leq y$
\item [2)] For all $x\in {\cal D}$, $\exists \bm{u}$ such that $$\Delta h(\bm x, \bm u) \geq  -\alpha(h(\bm x)),$$ 
where
$$\Delta h(\bm x, \bm u):= h\circ f(\bm{x},\bm{u}) - h(\bm{x}). $$
\end{itemize}
\end{definition}

\begin{remark}
Note that the safe set $\mathcal{S}$ is attractive if $h$ is a DCBF. This is seen by constructing a Discrete-time Control Lyapunov Function \cite{rssDCBF}: $$
V(\bm{x})=\left\{\begin{array}{ccc}
0, & \text { if } & \bm{x} \in \mathcal{S} \\
-h(\bm{x})\ge 0, & \text { if } & \bm{x} \in \mathcal{D} \backslash \mathcal{S}
\end{array}\right.
$$
such that we have:
\begin{equation*}
    \begin{aligned}
  & \Delta {V}(\bm{x}_k,\bm{u}_k)=V \circ f({{\bm{x}}_{k}},{{\bm{u}}_{k}})-V({{\bm{x}}_{k}}) \\ 
 & =-\Delta h({{\bm{x}}_{k}},{{\bm{u}}_{k}}) \\ 
 & \le \alpha (h({{\bm{x}}_{k}}))=\alpha (-V({{\bm{x}}_{k}}))\le 0  .
\end{aligned}
\end{equation*}
\end{remark}

\begin{remark}
Suppose {\color{black}$\alpha \in \mathcal{K}_{\infty,e}$} is a linear function such that $\alpha(t)=\gamma t, 0 < \gamma \leq 1$. If $h(\bm x_0)\geq0$, we have $h(\bm{x}_{k+1})\geq (1-\gamma)h(\bm{x}_{k})$. Furthermore, we have $h(\bm x_k) \geq (1-\gamma)^{k}h(\bm x_0)\geq0$, which is an exponential function in $k$, thus named Discrete-time Exponential Control Barrier Function~\cite{rssDCBF}. If $h(\bm{x}_0) < 0$, we will have $h(\bm x_k) \geq (1-\gamma)^{k}h(\bm x_0)$, which means the safe set $\mathcal{S}$ is also exponentially stable.
\end{remark}

\subsection{Model Predictive Control}

MPC is designed to resolve a constrained optimal control problem over a finite time horizon. \citet{Bike-MPC-DCBF} combine DCBF and MPC to achieve safe optimal performance for discrete-time systems. In our application, we utilize DCBF-MPC to design a feasible stride-to-stride motion that is energetically optimal with respect to a moving time horizon of fixed length. 
Compared to the use of distance constraints in MPC, the CBF constraints are more robust.

Given an initial condition ${\bm{x}}_{init}$ for the system at time $0$ and a time horizon $N$, the DCBF-MPC formulates the following optimization problem. 
\begin{problem}[DCBF-MPC]
\label{prob:DCBF-MPC}
\begin{equation*}
    \begin{matrix}
   \underset{{{({{\bm{X}}},{{\bm{U}}})}}}{\mathop{\min }}\,{{c}_{N}}({{\bm{x}}_{N}})+\sum\limits_{k=0}^{N-1}{{{c}_{k}}({{\bm{x}}_{k}},{{\bm{u}}_{k}})}  \\
   \begin{aligned}
  & s.t. \\ 
 & {{\bm{x}}_{k+1}}=f({{\bm{x}}_{k}},{{\bm{u}}_{k}}) \\ 
 & {{\bm{x}}_{0}}={{\bm{x}}_{init}}, {{\bm{x}}_{k}}\in \mathcal{D}, 
 {{\bm{u}}_{k}}\in \mathcal{U} \\ 
 & \Delta h({{\bm{x}}_{k}},{{\bm{u}}_{k}})\geq -\alpha(h({{\bm{x}}_{k}})), \\
\end{aligned}  \\
\end{matrix}
\end{equation*}
where $\bm{X}:=(\bm{x}_1,\bm{x}_2,\dots,\bm{x}_N), \bm{U}:=(\bm{u}_0,\bm{u}_1,\dots,\bm{u}_{N-1})$ and the running cost is $c_k, k=1,2,\dots,N$. 
\end{problem}

We redefine the legged robot system state as $\bm x_k = [x_k, \dot{x}_k, y_k, \dot{y}_k]^{\transpose}, \bm{u}=[p_{x,k}, p_{y,k}]^{\transpose}$. The system matrix for the new states are represented by $\bm{A}_f$ and $\bm{B}_f$ by reorganizing \eqref{eq:x_dyn}-\eqref{eq:y_dyn}.
For our application in bipedal robot motion planning, where we seek to steer the robot to a desired position subject to the safety and kinematic constraints given by \eqref{eq:leg_reachable_region}-\eqref{eq:minmax_length}, the problem is formulated as follows.
\begin{problem}[Safety-critical Multi-step Planner]
\label{prob:Biped-DCBF-MPC}
\begin{equation}
\label{eq:Biped-DCBF-MPC}
    \begin{matrix}
   \underset{{{\left( {{\bm{X}}},{{\bm{U}}} \right)}}}{\mathop{\min }}\,c_{N}(\bm{x}_N)  \\
   \begin{aligned}
  & s.t. \\ 
 & {{\bm{x}}_{k+1}}={{\bm{A}}_{f}}{{\bm{x}}_{k}}+{{\bm{B}}_{f}}{{\bm{u}}_{k}},{{\bm{x}}_{0}}={{\bm{x}}_{init}} \\ 
 & l{{b}_{x,k}}\le \sin ({{\theta }_{k}}){{p}_{x,k}}+\cos ({{\theta }_{k}}){{p}_{y,k}}\le u{{b}_{x,k}}, \\ 
 & l{{b}_{y,k}}\le -\cos ({{\theta }_{k}}){{p}_{x,k}}+\sin ({{\theta }_{k}}){{p}_{y,k}}\le u{{b}_{y,k}}, \\ 
 & {{l}_{\min }^2}\le \Delta x_{k}^{2}+\Delta y_{k}^{2}\le {{l}_{\max }^2}, \\ 
 & h(\bm{x}_{k+1})\geq (1-\gamma)h(\bm{x}_{k}). \\ 
\end{aligned}  \\
\end{matrix}
\end{equation}
Note that we assume the robot state starts inside the safe set, i.e., $h(\bm{x}_0)\geq0$. The CBF is not an explicit function of $\bm{u}$ as it is imposed by the constraints. 
\end{problem}

\begin{figure*}[t]
    \centering
   \includegraphics[width=1.5\columnwidth]{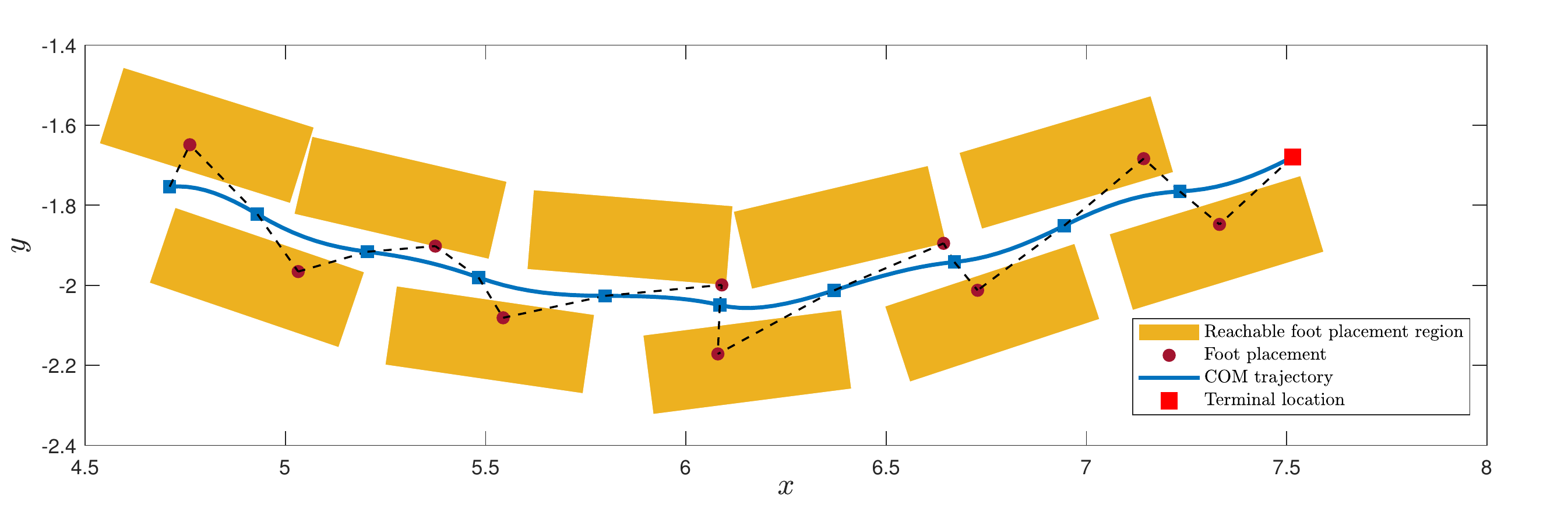}
    \caption{Path planned by simplified model. The reachable region of foot placement are the squares. The reachable regions are conservative approximations so the foot placements near the boundary are still feasible.}  \label{fig:reachable}
\end{figure*}

\begin{figure}
    \centering
   \includegraphics[width=0.99\columnwidth]{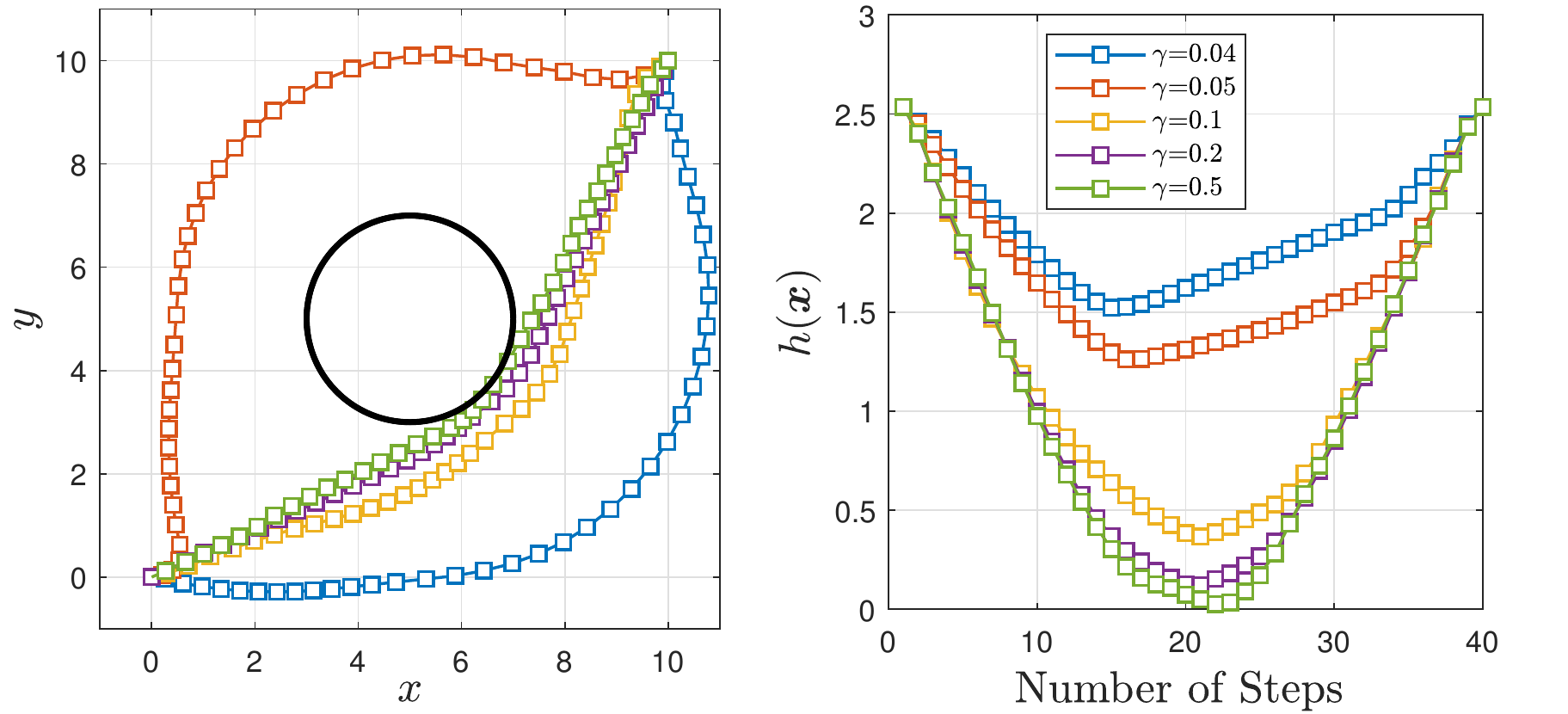}
    \caption{Trajectory optimization by DCBF-MPC on the LIP-based simplified model with different values of $\gamma$. The step number is  $N=40$. } 
    \label{fig:dcbf-mpc-gamma}
\end{figure}

The terminal cost function is designed to minimize the distance between the terminal positions and goal position $[x_f,y_f]^{\transpose}$ while regulating the terminal velocity to 0,
\begin{equation*}
    c_{N}(\bm{x}_N):={{w}_{1}}(\dot{x}_{N}^{2}+\dot{y}_{N}^{2})+{{w}_{2}}( {{({{x}_{N}}-{{x}_{f}})}^{2}}+{{({{y}_{N}}-{{y}_{f}})}^{2}}),
\end{equation*}
where $w_1$ and $w_2$ are positive weights. 
Part of a path planned by the DCBF-MPC and the associated foot placements are illustrated in Fig. \ref{fig:reachable}. We also compared the path plans with different $\gamma$ in Fig. \ref{fig:dcbf-mpc-gamma}. The robot starts from $(0\ m,0\ m)$ and the goal position is $(10\ m,10\ m)$. The obstacle is modeled as a circle centered at $(5\ m,5\ m)$ with radius $2\ m$. The function $h(\bm{x})$ is defined as:
\begin{equation}
\label{eq:cbf-ellipse1}
h({{\bm{x}}_{k}})=\left({{\left(\frac{{{x}_{k}}-5}{2}\right)}^{2}} + {{\left(\frac{{{y}_{k}}-5}{2}\right)}^{2}}\right)^{\frac{1}{2}}-1.
\end{equation}
We see that a path planned with a smaller value of $0 < \gamma \le 1$ is more conservative and robust, while a larger $\gamma$ allows the robot to more closely approach the obstacle, though it still remains outside. The different decay rate of $h(\bm{x})$ can also be verified in Fig. \ref{fig:dcbf-mpc-gamma}. 

\subsection{Application in Sampling-Based Motion Planning}
Now we incorporate the DCBF-MPC into the sampling-based motion planning framework RRT~\cite{RRT}. The algorithmic implementation of RRT-DCBF-MPC is presented in Algorithm \ref{RRT-DCBF-MPC}. The functions used in the algorithm are explained as follows.
\begin{itemize}
    \item \texttt{Sample} -- This function returns i.i.d. samples from the free space $\mathcal{X}_f$.
    \item \texttt{Nearest} -- Given a graph $\mathcal{G}=(\mathcal{V},\mathcal{E})$, where $\mathcal{V} \subset \mathcal{X}_f$, and a query 
point $\boldsymbol x \in \mathcal{X}_f$, this function returns a vertex $v \in \mathcal{V}$ that has the ``smallest'' distance to the query point.
    \item \texttt{DCBF\_MPC} -- This function extends nodes towards newly sampled points and allows for constraints on motion of the legged robot. It returns the first step by solving Problem~\ref{prob:Biped-DCBF-MPC} and the flag denoting the feasibility of the solution. The flag is \textbf{true} if the solution is feasible, otherwise it is \textbf{false}.
    \item \texttt{NoCollision} -- Given two points $\boldsymbol x_a, \boldsymbol x_b \in \mathcal{X}_f$, this functions returns \textbf{true} if the line segment between $\boldsymbol x_a$ and $\boldsymbol x_b$ is collision-free and \textbf{false} otherwise.
    \item \texttt{ComputeSteps} -- This function returns an approximation of the time-horizon for solving \texttt{DCBF\_MPC} given two points $\boldsymbol x_a, \boldsymbol x_b \in \mathcal{X}_f$. 
\end{itemize}

The algorithm's main loop begins with the start node and generating random sample points $x_{rand}$ in $\mathcal{X}_f$. For each new point sampled, the algorithm finds the nearest node in $\mathcal{V}$ and steers to the point using $\texttt{DCBF\_MPC}$. Inside the $\texttt{DCBF\_MPC}$ function, we give an estimation of the time horizon $N$ through $\texttt{ComputeSteps}$. To ensure multi-step stability, we set a minimal step number $N_{\text{min}}$. A maximum time horizon $N_{\text{max}}$ is also set considering the computational budget. We only reserve the first step planned by the MPC, which is very similar to the applications of MPC in real-time control~\citep{bangura2014real,di2018dynamic,ding2019real,liao2020time,Grandia-RSS-20}. When a path is chosen after the planning is finished, we will decide the heading angle using \eqref{eq:heading_angle_constrain}.  

\begin{algorithm}[t]
\caption{\texttt{RRT-DCBF-MPC()}} %
\small{
\label{RRT-DCBF-MPC}
\begin{algorithmic}[1]
\Require Free space $\mathcal{X}_{f}$, start configuration $\bm{x}_{start}$;
\State // Initialize node list, edge list, and tree
\State $\mathcal{V}\leftarrow \{\bm{x}_{start}\}$; $\mathcal{E} \leftarrow \varnothing$
\While{true} %
    \State $\bm{x}_{rand} \leftarrow \texttt{Sample}(\mathcal{X}_f)$
    \State $\bm{x}_{nearest} \leftarrow \texttt{Nearest}(\bm{x}_{rand},\mathcal{V})$
    \State $\bm{x}_{new}$, isFeasible $\leftarrow \texttt{DCBF\_MPC}(\bm{x}_{nearest}, \bm{x}_{rand})$ 
    \If{$\texttt{NoCollision}( \bm{x}_{nearest}, \bm{x}_{new}, \mathcal{X}_{f})$ \textbf{and} isFeasible}
         \State  $\mathcal{E} \leftarrow \cup \{(\bm{x}_{nearest}, \bm{x}_{new})\}, \mathcal{V} \leftarrow \cup \{\bm{x}_{new}\}$
         \If{\texttt{ReachGoal}($\bm{x}_{new}, \bm{x}_{goal}$)}
            \State \textbf{break}
         \EndIf
    \EndIf
\EndWhile
\State \Return $\mathcal{T} = (\mathcal{V}, \mathcal{E})$ \\
\Function{\texttt{DCBF\_MPC}}{$\bm{x}_{start}, \bm{x}_{goal}$}
    \State $N \leftarrow \texttt{ComputeSteps}(\bm{x}_{nearest}, \bm{x}_{rand})$
    \State $N \leftarrow \min(N, N_{max})$, $N \leftarrow \max(N, N_{min})$
    \State $(\bm{x}_{1,2,...N}, \bm{u}_{0,1,2,...,N-1}$,isFeasible$)$ $\leftarrow \argmin$ Problem \ref{prob:Biped-DCBF-MPC}
    \State $\bm{x}_{new} \leftarrow (\bm{x}_{1},\bm{u}_0)$
    \State \Return $\bm{x}_{new}$, isFeasible
\EndFunction
\end{algorithmic}
}
\end{algorithm}

%

\section{Safe Informative Motion Planning}
\label{sec:safeiig}

\citet{RIG} formulated the Robotic Information Gathering (RIG) problem as a maximization problem subject to finite resources, i.e., a budget $b$. \citet{IIG} extended RIG to an incremental motion planning setup, IIG, via a convergence criterion based on Relative Information Contribution (RIC). This convergence criterion enables the robot to execute planned actions autonomously as opposed to being anytime (manual stop). Our proposed planner, SAFE-IIG, is built on the IIG algorithm~\citep{IIG}. 

\begin{algorithm}[t!]
\caption[SAFE-IIG]{\texttt{SAFE-IIG}()}
\small{
\label{alg:safe-iig}
\begin{algorithmic}[1]
\Require Budget $b$, free space $\mathcal{X}_{f}$, Environment $\mathcal{M}$, start configuration $\boldsymbol x_{start}$, near radius $r$, relative information contribution threshold $\delta_{RIC}$, averaging window size $n_{RIC}$;
\State // Initialize cost, information, starting node, node list, edge list, and tree
\State $I_{init} \gets \texttt{Information}([\ ],\boldsymbol x_{start},\mathcal{M}), C_{init} \gets 0, n \gets \langle \boldsymbol x_{start}, C_{init}, I_{init} \rangle$ \label{line:iig_inits}
\State $\mathcal{V} \gets \{n\}, \mathcal{V}_{closed} \gets \varnothing, \mathcal{E} \gets \varnothing$
\State $n_{sample} \gets 0$ // Number of samples  
\State $I_{RIC} \gets \varnothing$ // Relative information contribution \label{line:iig_inite}
\While {$\texttt{AverageRIC}(I_{RIC},n_{RIC}) > \delta_{RIC}$} $\label{iigcondition}$
\State // Sample configuration space of vehicle and find nearest node
\State $\boldsymbol x_{sample} \gets \texttt{Sample}(\mathcal{X}_f)$ \label{line:iig_samples}
\State $n_{sample} \gets n_{sample} + 1$
\State $\boldsymbol x_{nearest} \gets \texttt{Nearest}(\boldsymbol x_{sample}, \mathcal{V} \backslash \mathcal{V}_{closed})$
\State $\boldsymbol x_{feasible},$isFeasible $ \gets \texttt{DCBF\_MPC}(x_{nearest}, x_{rand})$ 
\If{not  isFeasible} 
\State \textbf{continue}
\EndIf \label{line:iig_samplee}
\State // Find near points to be extended
\State $\mathcal{V}_{near} \gets \texttt{Near}(\boldsymbol x_{feasible}, \mathcal{V} \backslash \mathcal{V}_{closed}, r)$ \label{line:iig_ballr}
\For{all $n_{near} \in \mathcal{V}_{near}$}
\State // Extend towards new point
\State $\boldsymbol x_{new},\ $isFeasible $ \gets \texttt{DCBF\_MPC}(x_{nearest}, x_{rand})$ \label{line:iig_newnode}
\If{$\texttt{NoCollision}(\boldsymbol x_{near}, \boldsymbol x_{new}, \mathcal{X}_{f})$ \textbf{and} isFeasible}  \label{line:iig_colfs}
\State // Calculate new information and cost
\State $I_{new} \gets \texttt{Information}(I_{near},\boldsymbol x_{new},\mathcal{M})$ 
\State $c(\boldsymbol x_{new}) \gets \texttt{Cost}(\boldsymbol x_{near}, \boldsymbol x_{new})$
\State $C_{new} \gets C_{near} + c(\boldsymbol x_{new})$
\State $n_{new} \gets \langle \boldsymbol x_{new}, C_{new}, I_{new} \rangle$ \label{line:iig_colfe}
\If{$\texttt{Prune}(n_{new})$}  \label{line:iig_prunes}
\State \textbf{delete} $n_{new}$
\Else
\State $I_{RIC} \gets \texttt{append}(I_{RIC}, (\frac{I_{new}}{I_{near}} - 1)/n_{sample})$ \label{line:iig_ric} 
\State $n_{sample} \gets 0$ // Reset sample counter
\State // Add edges and nodes
\State $\mathcal{E} \gets \cup \{(n_{near}, n_{new})\}, \mathcal{V} \gets \cup \{n_{new}\}$
\State // Add to closed list if budget exceeded
\If{$C_{new} > b$}
\State $\mathcal{V}_{closed} \gets \mathcal{V}_{closed} \cup \{n_{new}\}$ \label{line:iig_prunee}
\EndIf
\EndIf
\EndIf
\EndFor
\EndWhile \\
\Return $\mathcal{T} = (\mathcal{V}, \mathcal{E})$
\end{algorithmic}}
\end{algorithm}


The algorithmic implementation of SAFE-IIG is shown in Algorithm~\ref{alg:safe-iig}. 

\begin{itemize}
    \item \texttt{Near} -- Given a graph $\mathcal{G}=(\mathcal{V},\mathcal{E})$, where $\mathcal{V} \subset \mathcal{X}_f$, a query 
    point $\boldsymbol x \in \mathcal{X}_f$, and a positive real number $r \in \mathbb{R}_{>0}$, this function returns a set of vertices $\mathcal{V}_{near} \subseteq \mathcal{V}$ that are contained in a ball of radius $r$ centered at $\boldsymbol x$.
    \item \texttt{Information} -- This function quantifies the information quality of a collision-free path between two points from the free space $\mathcal{X}_f$. In this work, we use the mutual information between the map and a depth camera~\citep[Algorithm 3]{IIG}.
    \item \texttt{Cost} -- The cost function assigns a strictly positive cost to a collision-free path between two points from the free space $\mathcal{X}_f$.
    \item \texttt{Prune} -- This function implements a pruning strategy to remove nodes that are not ``promising''. This can be achieved through defining a \emph{partial ordering} for co-located nodes.
\end{itemize}

Lines~\ref{line:iig_inits}-\ref{line:iig_inite} show the algorithm initialization. 
In lines~\ref{line:iig_samples}-\ref{line:iig_samplee}, a feasible sample point from $\mathcal{X}_f$ is drawn. Line~\ref{line:iig_ballr} extracts all nodes from the graph that are within radius $r$ of the feasible point. These nodes are candidates for extending the graph, and each node is converted to a new node using the \texttt{DCBF\_MPC} function in line~\ref{line:iig_newnode}. In lines~\ref{line:iig_colfs}-\ref{line:iig_colfe}, if there exists a collision free path between the candidate node and the new node, the information gain and cost of the new node are evaluated. In lines~\ref{line:iig_prunes}-\ref{line:iig_prunee}, if the new node does not satisfy a partial ordering condition it is pruned, otherwise it is added to the graph. Furthermore, the algorithm checks for budget constraint violation. The output is a graph that contains a subset of safe and dynamically feasible paths with maximum information gain.
\section{Simulation Results on a High-Dimensional 3D Biped Robot}
\label{sec:results}
\begin{figure}[t]
    \centering
  \includegraphics[width=0.99\columnwidth]{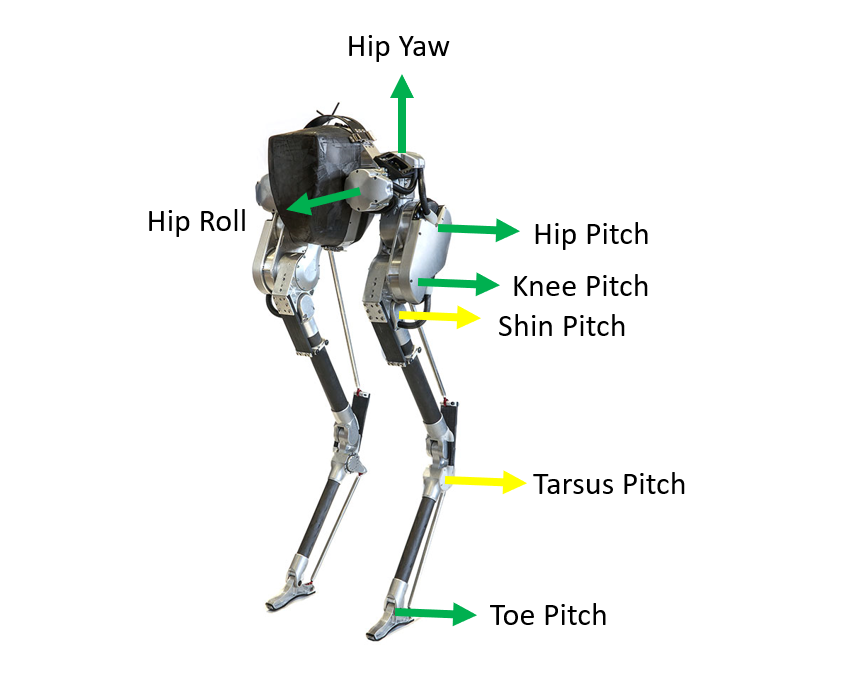}
    \caption{Cassie by Agility Robotics is a 3D robot with seven joints on each leg, where five are actuated by motors and two are constrained by springs.}
    \label{fig:cassie}
\end{figure}

We apply the proposed algorithms to motion planning of a 20 degree of freedom bipedal Cassie-series robot, shown in Fig.~\ref{fig:cassie}, and validate the feasibility of the planned path in simulation. The Cassie robot's weight is $32\ \mathrm{kg}$, and its CoM height is $H=0.6 \m$. The limits of the reachable region are $ub_{x_b}=0.3 \m, lb_{x_b}=-0.2 \m$ in the robot's sagittal plane for all steps. When the next step is right stance, the bounds in the frontal plane are $ub_{y_b}=0.25 \m, lb_{y_b}=0.05 \m$, and $ub_{y_b}=-0.05 \m, lb_{y_b}=-0.25 \m$ for left stance.  

We first present the path planned by RRT-DCBF-MPC in a single obstacle environment. Next, we present the results of SAFE-IIG in a dense stochastic map. We then introduce  LIP model-based bipedal walking controller \cite{yukaiAngularMomentum} for robot control to track the path planned by the proposed algorithms.

The simulations are launched in Matlab Simulink. We use the Matlab function $\texttt{fmincon}$ with interior point method to solve Problem \ref{prob:Biped-DCBF-MPC} for each sample. We use $N_{max} = 3$ as the maximal time horizon and $N_{min} = 2$ as the minimum. The 3-step planning is long enough to provide foresight to the future while limiting the computational burden. Without analytical gradients, the optimization can run at 5 Hz on a laptop equipped with a 4-core Intel i7-7820HQ 2.9 GHz CPU.

\subsection{Single obstacle environment} 
We apply RRT-DCBF-MPC to legged robot path planning in a $25 \m \times 25 \m$ square region. We consider an ellipsoidal obstacle centered at $(10 \m, 10 \m)$ with long axis being $8 \m$ and short axis being $1 \m$. We choose $\gamma = 0.75$ in this numerical simulation. Similar to \citep{rssDCBF}, the CBF constraints are
\begin{equation}
\label{eq:cbf-ellipse2}
h({{\bm{x}}_{k}})={{({{x}_{k}}-10)}^{2}} + {{(\frac{{{y}_{k}}-10}{8})}^{2}}-1 .
\end{equation}

\begin{figure}
    \centering
  \includegraphics[width=0.75\columnwidth]{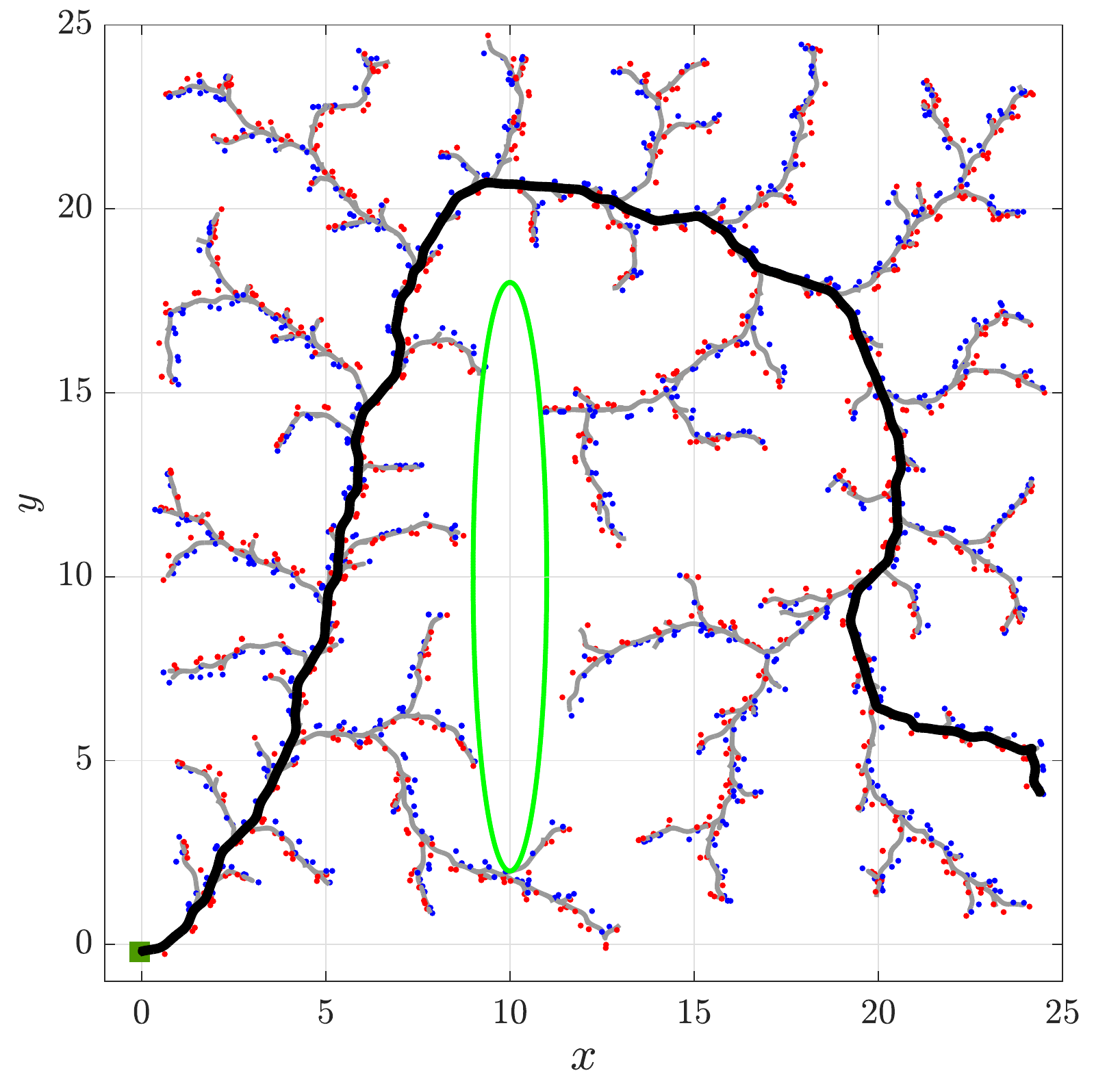}
    \caption{Legged robot motion planning by RRT-DCBF-MPC in a single obstacle environment. The CoM trajectory is plotted in grey. Blue and red dot denotes the left and right foot placement, respectively. We randomly choose a path for tracking, which is plotted in black.}
    \label{fig:rrt-single}
\end{figure}

As the RRT-based algorithm can be stopped anytime and we do not specify a target region, we sampled 2500 points in the $x-y$ plane to extend the tree. The planned path, including the CoM trajectory and foot placements, is presented in Fig. \ref{fig:rrt-single}. The robot's trajectory successfully avoids the obstacle. Though the path is not smooth, we will later show that it is dynamically feasible and can be tracked by the controller.

\subsection{Stochastic Dense Map}
We now implement the SAFE-IIG in the Cave map~\citep{Radish_data_set}. The map is initialized as an occupancy grid map by assigning each point an occupied or unoccupied probability. We model a non-uniformly distributed signal by placing two signal sources at the top of the map. The strength of the two signals decays exponentially with distance:
\begin{equation*}
p(\bm{x})=\sum\limits_{n}{{{\lambda }_{n}}\exp (-{{\left\| \bm{x}-{{\bm{x}}_{n}} \right\|}_{{{\Sigma }_{n}}}})}, \quad \bm{x}\in {{\mathcal{X}}_{f}}
\end{equation*}
where $\bm{x}_n$ is the center of the signal source, and $\lambda_n$ is the signal strength. Here we use Mahalanobis distance ${{\left\| \cdot \right\|}_{{{\Sigma }_{n}}}}$.

To define the safe set, we first use a clustering method, e.g., k-means to group the occupied points. We then use $p$-norm balls to approximate each detected obstacle and define the CBF $h({{\bm{x}}_{k}})={{\left\| \bm\Sigma ({{\bm{r}}_{k}}-{{\bm{r}}_{obs}}) \right\|}_{p}}-1\geq 0$,
where $\bm{r}_k := [x_k,y_k]^{\transpose}$ represents the robot's Cartesian position at step $k$ and $\bm{r}_{obs}$ denotes the center of an obstacle. $\bm\Sigma$ is a linear transformation that rotates and normalizes the axis. As DCBF allows the robot state to approach the boundary of safe set, we add a buffer to the radius of each ball. For example, the obstacle at the top of the map is written as:
\begin{equation}
\label{eq:cbf-p-norm}
    h({\bm{x}_{k}})={{\left( {{\left| \frac{{{x}_{k}}-{{x}_{c}}}{{{r}_{x}}+{{r}_{buff,x}}} \right|}^{p}}+{{\left| \frac{{{y}_{k}}-{{y}_{c}}}{{{r}_{y}}+{{r}_{buff,y}}} \right|}^{p}} \right)}^{\frac{1}{p}}}-1,
\end{equation}
where the parameters $x_c=4.2, y_c=18.9$ are the center, $r_x = 1.9, r_y = 0.9$ the radii, and $r_{buff, x} = r_{buff,y}=0.5$ the buffer size. We choose $p = 10$ to approximate the infinity norm while avoiding the non-smoothness. When applied in a map, only the obstacles near $\bm{x}_{nearest}$ will activate the corresponding CBF constraint for the MPC. 

\begin{figure}[t]
    \centering
  \subfloat{\includegraphics[width=0.75\columnwidth]{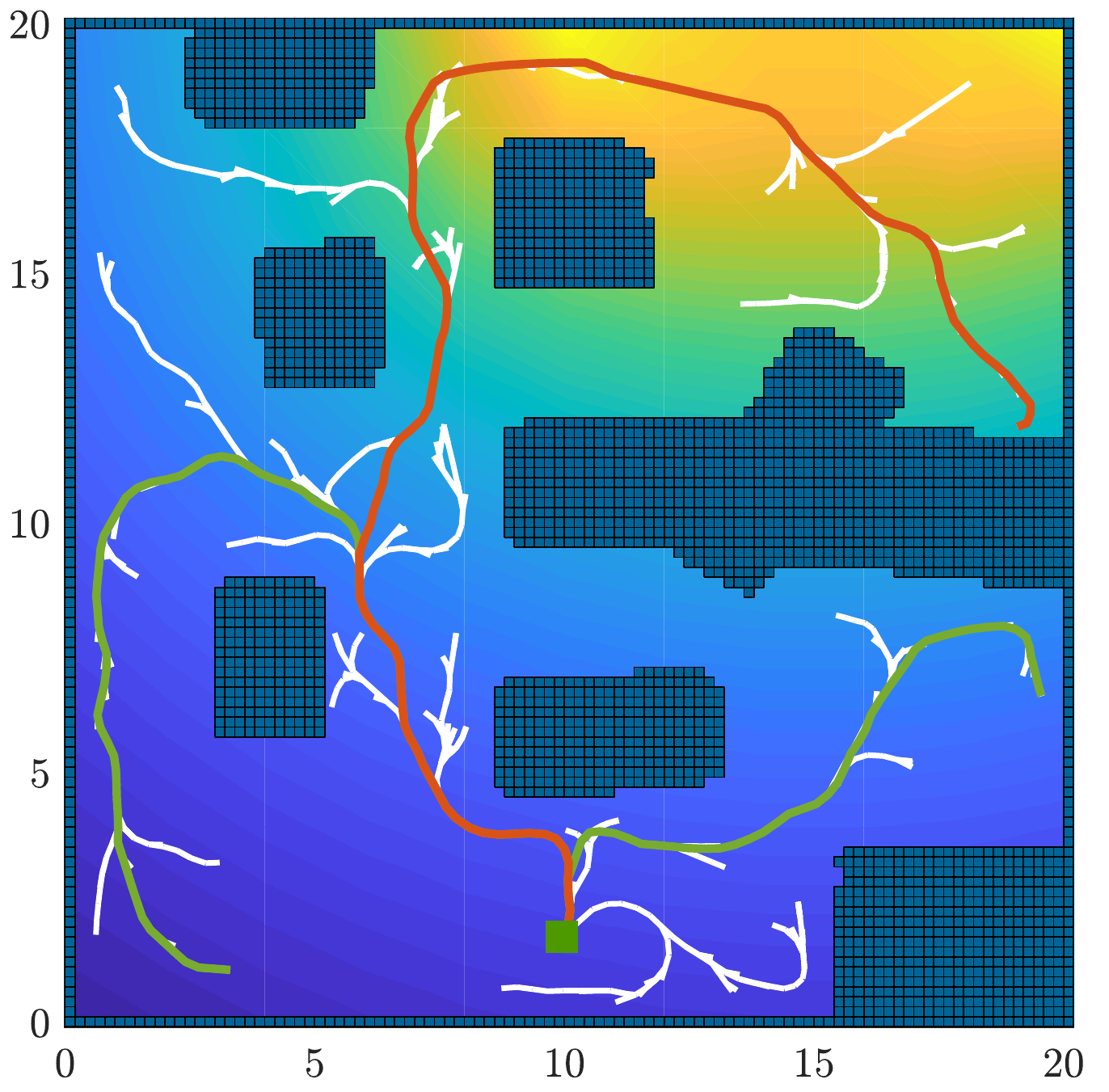}}\\
  \subfloat{\includegraphics[width=0.75\columnwidth]{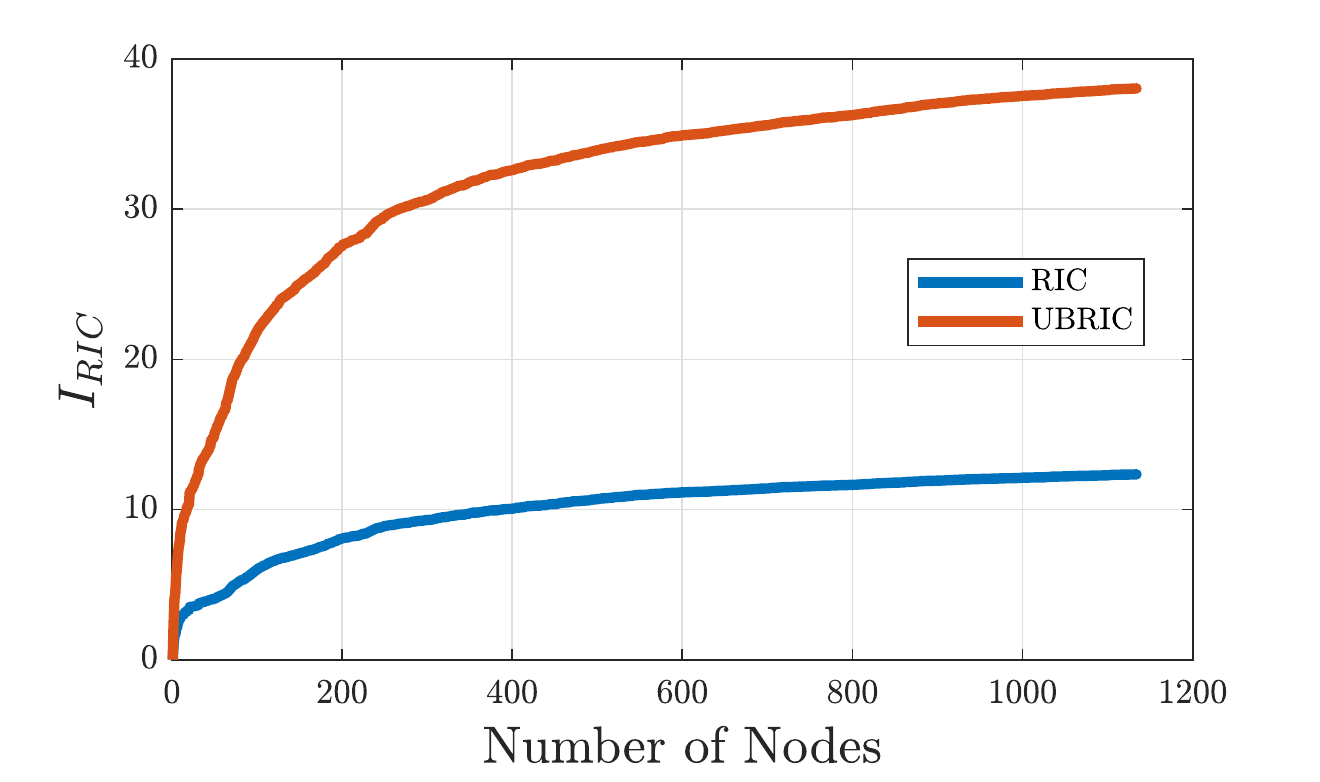}}
    \caption{Top: Path plan by SAFE-IIG in cave map. The red path maximizes the information gathered. Bottom: Convergence graph of the cumulative penalized $I_{RIC}$. The planning converges automatically when RIC is below the threshold.}
    \label{fig:iig}
\end{figure}

The planning stops when the Relative Information Contribution (RIC)~\citep{IIG} is below the threshold $\delta_{RIC}$ (e.g., $5e-3$). RIC is essentially the non-dimensional information gain and shows the contribution of any new node in the graph relative to its parent node. The final path plans are presented in Fig.~\ref{fig:iig}. The signal distributions are represented by the color in the unoccupied region, where a brighter color denotes a stronger signal. The planned path is in white. Three paths are selected and the path with maximized information gain is plotted in red. The evolution of RIC and its upper bound (UBRIC) are presented in Fig.~\ref{fig:iig}. The upper bound is based on the map entropy rather than the mutual information between the map and measurements from a simulated depth camera~\citep{IIG}.

We observe that the planned path maintains a clear distance from the obstacles. As multiple CBF constraints can be activated at the same time in this map, the robot path is constrained in a safe corridor and thus is much smoother than in the RRT-DCBF-MPC case. In the following section, we will show that the planned path is also dynamically feasible by simulation.

\subsection{LIP-based Angular Momentum Controller}


The controller is developed in \cite{yukaiAngularMomentum}. The linear velocity term is replaced with angular momentum about the contact point $\bm{L}$. On a real robot, $\bm{L} = \bm{L}_{\rm CoM} + \bm{p} \times m\bm{v}_{\rm CoM}$, where $\bm{L}_{\rm CoM}$ is the angular momentum about the CoM, $\bm{p}$ is the vector from the contact point to the CoM. We neglect $\bm{L}_{\rm CoM}$ for walking task. For LIP model, ${L}^y = -mHp_x$ and the dynamics is
\begin{equation} \label{eq:LIP_AM_eqn}
\begin{bmatrix}
\dot{x} \\
\dot{L}^y \\
\end{bmatrix}
=
\begin{bmatrix}
0 & \frac{1}{mH} \\
0 & 0\\
\end{bmatrix}
\begin{bmatrix}
x \\
L^y
\end{bmatrix}
- 
\begin{bmatrix}
0 \\
mg
\end{bmatrix}
p_x
\end{equation}

Compared to the original LIP model, this model has a higher fidelity to the real robot model because of several desirable properties of $\bm{L}$ as discussed by \citet{yukaiAngularMomentum}.

The dynamics from step to step can be described by:
\begin{equation*}
    \left[ \begin{aligned}
  & {{x}_{k+1}} \\ 
 & L_{k+1}^{y} \\ 
\end{aligned} \right]={{\bm{A}}_{L}}\left[ \begin{matrix}
   {{x}_{k}}  \\
   L_{k}^{y}  \\
\end{matrix} \right]+{{\bm{B}}_{L}}{{p}_{x,k}}
\end{equation*}
\begin{equation*}
\small{
{{\bm{A}}_{L}}:=\left[ \begin{matrix}
   1 & \tfrac{1}{mH\beta }\sinh (\beta T)  \\
   0 & \cosh (\beta T)  \\
\end{matrix} \right],{{\bm{B}}_{L}}:=\left[ \begin{matrix}
   1-\cosh (\beta T)  \\
   -mH\beta \sinh (\beta T)  \\
\end{matrix} \right]
}
\end{equation*}
Substitute desired angular momentum to the left side of \eqref{eq:LIP_AM_eqn}, we can decide the desired foot placement:
\begin{equation} \label{eq:Desired_Footplacement}
    p_{x,k} =\frac{-L^{y ~{\rm des}}_{k+1} + \cosh(\beta T)L^{y}_{k}}{mH \beta \sinh(\beta T)}.
\end{equation}
On Cassie robot, after deciding the foot placement, time-based reference trajectories are generated for the following nine control variables.
\begin{equation*} 
\label{eq:control variables}
\footnotesize{
z_0=
\begin{bmatrix}
\rm torso \; pitch \\
\rm torso \; roll \\
\rm stance \; hip \; yaw \\
\rm swing \; hip \; yaw \\
\rm CoM\ height \\
p^x_{\rm swing foot} \\
p^y_{\rm swing foot} \\
p^z_{\rm swing foot} \\
\rm swing \;toe \;absolute \;pitch \\
\end{bmatrix}.
}
\end{equation*}

The reference for torso pitch and roll are constant 0 to keep the torso upright. Hip yaw on two legs changes linearly w.r.t. time to track the heading angle. CoM height is kept constant to imitate LIP model. Swing foot position will arrived at the decided foot placement at the end of a step. Swing toe pitch is constant zero to keep the foot flat.

After the reference trajectories are generated, a low-level Input-Output Linearization controller is implemented to make control variables track their references. An Input-Output Linearization controller enforces the following \textit{linear} equation through \textit{input}:
\begin{equation*}
    \ddot{z} + K_p\dot{z} + K_d z = 0,
\end{equation*}
where $z = z_0 - z_d$ is the so-called \textit{output}, $z_d$ is the reference for the control variables. $K_p$ and $K_d$ are chosen such that eigenvalues are negative.
More details about the advantage of choosing angular momentum as state variable, selection of output states, and the implementation on a real robot are described in \cite{yukaiAngularMomentum}.

\subsection{Path Tracking}
We use the LIP-based angular momentum controller to track the desired paths. We consider 2 scenarios that interests us. The first scenario is called open-loop tracking, where we assumes no position feedback. The second scenario is closed-loop tracking, such that position feedback is available. In both scenarios we assume we know the robot's yaw angle as a reliable estimate is accessible through the IMU when the initial value is known. 

In the open-loop tracking, we use the angular momentum at the waypoint and the heading yaw angle as the feedback. The heading angle is set to be the desired angle of the swing toe.
We plug the angular momentum obtained by the planner at waypoint $k+1$ as desired angular momentum into \eqref{eq:Desired_Footplacement} to get the foot placement $p_{x,k}$.

In the closed-loop tracking, CoM position and angular momentum are simultaneously controlled. Similar to the deadbeat control described by~\citet{Xiong_2020}, foot placement for $k$ and $k+1$ steps are calculated to obtain desired position and angular momentum at $k+2$ step, using
\begin{equation}
\label{eq:deadbeat}
    \left[ \begin{aligned}
  & {{x}_{k+2}} \\ 
 & L_{k+2}^{y} \\ 
\end{aligned} \right]={{\bm{A}}_{L}}^2\left[ \begin{matrix}
   {{x}_{k}}  \\
   L_{k}^{y}  \\
\end{matrix} \right]+{{\bm{A}}_{L}}{{\bm{B}}_{L}}{{p}_{x,k}} + {{\bm{B}}_{L}}{{p}_{x,k+1}} .
\end{equation}
Given the state at step $k$ and the desired state at step $k+2$, we can solve \eqref{eq:deadbeat} for foot placement $p_{x,k}$ and $p_{x,k+1}$.
Only foot placement $p_{x,k}$ is implemented.

\begin{figure*}[t]
    \centering
  \includegraphics[width=1.5\columnwidth]{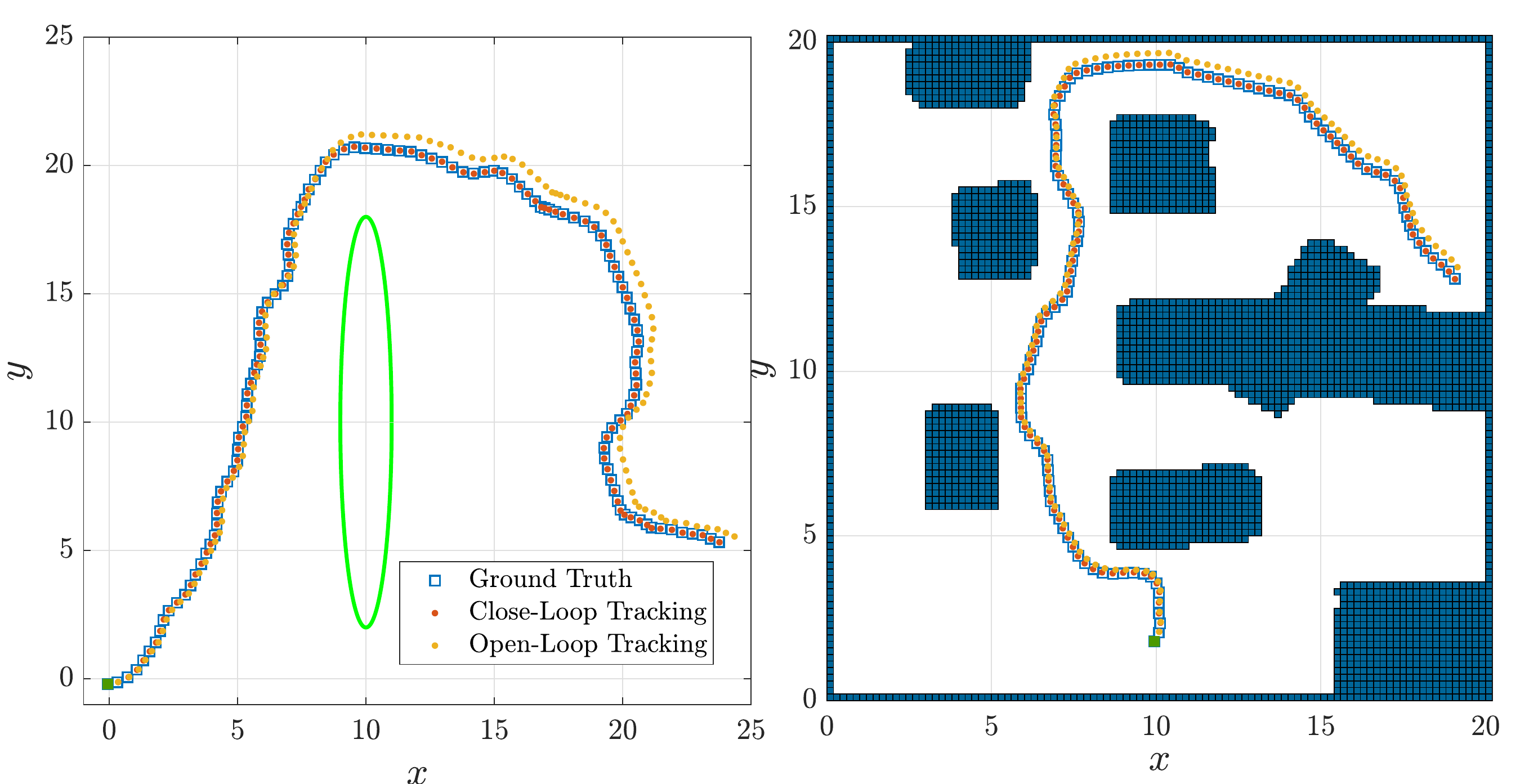}
    \caption{Robot CoM trajectory in the simulation. Left: path planned by RRT-DCBF-MCP. Right: path planned by SAFE-IIG. Compared to RRT-DCBF-MCP, the computed path by SAFE-IIG is smoother, and the drifting in open-loop tracking is smaller. No re-planning is involved in the simulation.}
    \label{fig:path-track}
\end{figure*}

\begin{figure}[t]
    \centering
   \includegraphics[width=0.99\columnwidth]{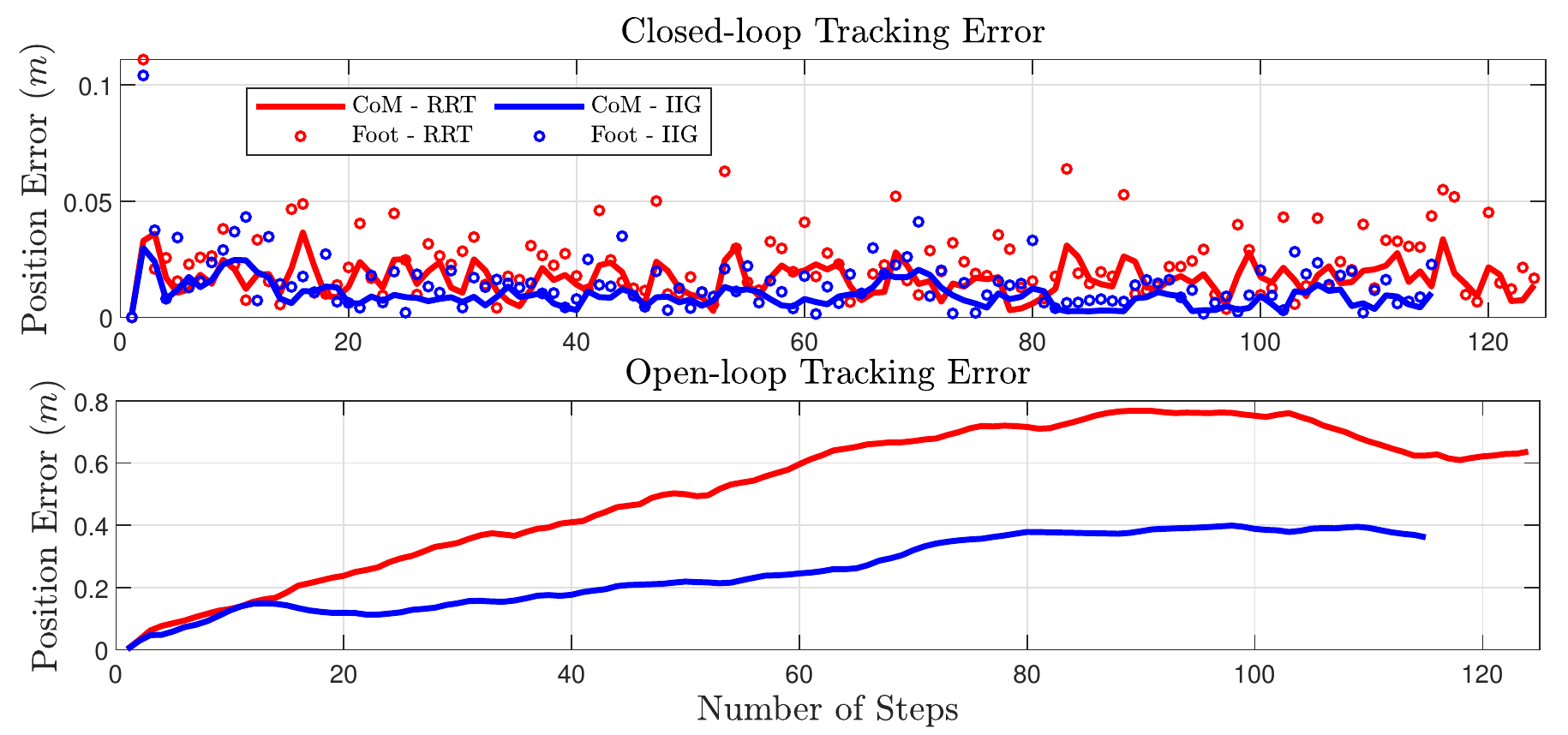}
    \caption{Error of path tracking. The CoM position error for closed-loop tracking are smaller than $0.05 \m$ and the foot placement error are less than $0.1 \m$ for both paths. In open-loop tracking, the error of CoM position is less than $0.8 \m$ in the single obstacle environment with 124 steps and less than $0.4 \m$ in cave map with 115 steps.} 
    \label{fig:path-track-error}
\end{figure}

Figure~\ref{fig:path-track} shows the simulated robot trajectories. The waypoint tracking error is presented in Fig.~\ref{fig:path-track-error}. The closed-loop tracking is close to the ground truth, while in open-loop tracking, the robot trajectories gradually drift from the desired path. However, the drift is small considering the path length. 

The results suggest that the planned paths are dynamically feasible considering the discrepancy between the simulated paths and desired paths. Additionally, we have demonstrated that the heading angle and the position and velocity at the waypoints can be directly used as commands for the low-level controller. 

\section{Discussions and Limitations}
\label{sec:discussion}
We combined DCBF-MPC with sampling-based  motion planning algorithms to plan safe dynamically feasible paths. For constructions of safe sets, the obstacle \eqref{eq:cbf-ellipse2} can be considered as the form $h({{\bm{x}}_{k}})={{\left\| \bm\Sigma ({{\bm{r}}_{k}}-{{\bm{r}}_{obs}}) \right\|}^{p}_{p}}-1$. Although this form works well when $p=2$, we observed that when $p$ becomes large, the CBF defined in \eqref{eq:cbf-p-norm} scales better when applied in the optimization. The planned path is not smooth due to random sampling. However, we can use it as an initial guess to obtain a smoother path via optimization, see the video in the supplementary material. 

We assumed full knowledge of the robot's dynamics model when using the Input-Output Linearization controller, and the state feedback is deterministic. An interesting future study for transferring this work on hardware is to fuse inertial and kinematics data to estimate the robot states~\citep{hartley2020contact} and apply a more robust controller to modeling error, e.g., \cite{yukaiAngularMomentum}. Moreover, we considered flat ground and neglected the vertical motion of the robot. In the low-level control part, the controller also does not consider safety criteria. In future work, we intend to integrate vertical motion in order to traverse complex terrains. In the context of state estimation, this challenge is also discussed by~\citet[see Figure 8]{hartley2018hybrid}. DCBF-MPC can also be used for real-time re-planning when a disturbance is involved. 


\section{Conclusion}
\label{sec:conclusion}
We developed an integrated framework for safety-aware informative motion planning, SAFE-IIG, suitable for legged robots. We integrated the Discrete-time Control Barrier Function and Model Predictive Control into the sampling-based motion planning frameworks, enabling safety-critical multi-step planning for legged robots. The DCBF-MPC plans a multi-step trajectory in each sampling, and it uses the first step for node expansion. SAFE-IIG can plan a collision-free path for legged robots while maximizing the information gathered along the path. The simulation results show that SAFE-IIG can plan a safe and dynamically feasible path while exploring a dense map.

In the future, we plan to implement SAFE-IIG on hardware for real-time exploration and mapping using biped and quadruped robots in unknown unstructured environments. 

\section*{Acknowledgments}
Toyota Research Institute (TRI) provided funds to support this work.  Funding for J. Grizzle was in part provided by TRI and in part by NSF Award No. 1808051.
\small 
\bibliographystyle{unsrtnat}
\bibliography{strings-full,ieee-full,references}

\end{document}